# Unsupervised Protoform Reconstruction through Parsimonious Rule-guided Heuristics and Evolutionary Search

Promise Dodzi Kpoglu
*LLACAN, CNRS*

**Abstract**

We propose an unsupervised method for the reconstruction of protoforms i.e., ancestral word forms from which modern language forms are derived. While prior work has primarily relied on probabilistic models of phonological edits to infer protoforms from cognate sets, such approaches are limited by their predominantly data-driven nature. In contrast, our model integrates data-driven inference with rule-based heuristics within an evolutionary optimization framework. This hybrid approach leverages on both statistical patterns and linguistically motivated constraints to guide the reconstruction process. We evaluate our method on the task of reconstructing Latin protoforms using a dataset of cognates from five Romance languages. Experimental results demonstrate substantial improvements over established baselines across both character-level accuracy and phonological plausibility metrics.

## 1. Introduction

The study of the evolution of languages is built on the assumption that today's languages are descendants of earlier ancestral languages, which are commonly referred to as *protolanguages*. The goal of the domain of linguistics that deals with this line of investigation, historical linguistics, is therefore among others, to uncover the phonological, lexical, and grammatical characteristics of these protolanguages and to understand the processes through which they evolved into present-day linguistic systems. Central to this endeavor is the *comparative method*, a well-established analytical procedure that has been the cornerstone of historical linguistic research for over a century (Campbell, 2013).

The comparative method involves a sequence of interrelated steps that include establishing language relatedness hypotheses, collecting and curating wordlists, identifying cognate sets (lexical items across languages that derive from a common ancestral form), detecting regular sound correspondences, and ultimately reconstructing protoforms i.e. hypothetical ancestral words from which words in modern languages, often referred to as *reflexes*, have evolved (Weiss, 2015). For instance, an example involves the English word *father*, the German *vater*, the Greek *pater*, and the Sanskrit *pitṛ́*. All of these are reflexes of a reconstructed Proto-Indo-European (PIE) root *\*poter-*. More importantly, this reconstruction is grounded in consistent sound correspondences observed across Indo-European languages, such as the PIE phoneme *\*p* regularly transforming into /f/ in English. The identification and systematization of such correspondences are what allow historical linguists to postulate both the protoforms themselves and the diachronic phonological rules that account for their evolution into current forms.

Although traditional reconstructions have been largely manual and expert-driven, the increasing availability of digitized language resources has inspired computational efforts to automate the process. Early attempts at computational protoform reconstruction relied on rule-based combinatorial frameworks (Durham & Rogers, 1969; Lowe & Mazaudon, 1994). These methods, while pioneering, were constrained by their rigid architectures and limited ability to accommodate probabilistic uncertainty inherent in linguistic evolution. Recent approaches have largely shifted towards data-driven and statistically grounded methods, falling on probabilistic models and machine learning techniques to infer protoforms (Bouchard et al., 2007; Bouchard-Côté et al., 2009; He et al., 2022 , etc.). While these methods represent significant progress, they often disregard an important consensus in historical linguistics: that the design of plausible phonological systems should not be fully ceded to automated systems without expert oversight (cf. Kondrak 2002).

This paper proposes a hybrid framework that integrates the interpretability and structure of rule-based systems with the adaptive learning capabilities of statistical models. Specifically, we introduce a multi-phase protoform reconstruction pipeline that combines parsimony-based heuristics with rule-based transformation, and grounded in an evolutionary search algorithm. The proposed method begins with the generation of candidate protoforms using parsimony principles i.e., favoring forms that minimize the number of phonological changes required to account for observed reflexes. These candidate forms are reranked based on their explanatory adequacy with respect to the reflex data. The reranked candidate protoforms are then transformed through manually assembled phonological rules. The original reranked candidate protoforms and their rule-transformed counterparts are then combined into a single set. This hybrid seed set serves as the input to an evolutionary algorithm, which simulates a reverse-Darwinian selection process. In this simulation, protoforms evolve under selective pressures defined by both statistical likelihoods and rule-based constraints. The evolutionary framework thus allows for a flexible exploration of the solution space while retaining phonological plausibility.

| WORDLIST | | PROTOFORM RECONSTRUCTION | | | |
|---|---|---|---|---|---|
| Language | Cognates | Ranked Parsimony heuristics | Sample Rules | Ranked rule-based heuristics | Evolution |
| Romanian | -an | -anõ | ('a','ă') | -an | |
| French | - | -anõ | | | |
| Italian | -ano | -ãõ | ('aʊ', 'o'), ('ʊm', 'o'), ('ʊs', 'o'), ('a','a'), ('o', 'o') | -anʊm | -anʊm |
| | | -ɐnõ | | | |
| | | -ãõ | | | |
| Spanish | -ano | | ('aʊ', 'o'), ('ʊm', 'o'), ('ʊs', 'o'), ('ʊ', 'o') | -anʊm | |
| Portugeuse | -ẽõ | | ('o','o'), (ʊ, o) | -ẽõ | |

**Figure 1:** *A scenario illustrating how a purely data-driven approach can lead to a constrained reconstructed phonemic system, as demonstrated by the highest-ranked parsimonious reconstruction 'anõ'. Incorporating both the parsimonious reconstruction and language-specific ranked rule-based reconstructions into a probabilistic evolutionary algorithm enables expansion of the phonemic inventory. In this case, the highest-ranked rule-based reconstruction '-anʊm' is a rule-based transformation that ultimately emerges as the most likely reconstruction.*

This hybrid approach addresses a key limitation of prior models i.e., their tendency to be conservative in the reconstructed phonemic inventory. By integrating multiple sources of inductive bias (statistical and rule-based), the model expands the range of phonemes and phonotactic patterns it can consider, thereby enhancing its ability to recover linguistically plausible protoforms. Figure 1 illustrates an example where the inclusion of rule-transformed seeds enables the model to reconstruct phonemes (/ʊ/ and /m/) that are absent in the reflexes.

To evaluate the effectiveness of the proposed model, we conduct a series of ablation studies that test the contribution of each component (parsimony ranking, rule-based seeding, and evolutionary search) to overall performance. The proposed model is benchmarked against both a parsimony-only model and a baseline expectation-maximization-based model, using the Romance language dataset curated by Meloni et al., (2021). Experimental results indicate that the proposed hybrid system outperforms these baselines in both reconstruction accuracy and phonological coherence.

The contributions of this study are threefold:

1. Innovation: we present an unsupervised protoform reconstruction algorithm that integrates parsimony-based heuristics and rule-based transformation with evolutionary search strategies.
2. Methodological integration: The system bridges the gap between rule-based linguistic reasoning and statistical inference, providing a principled way to combine expert knowledge with data-driven discovery.
3. Scalability and Extensibility: The proposed architecture paves the way for future extensions that could incorporate sequence-to-sequence architectures, or attention-based models for further performance gains.

Thus, this paper contributes a novel, linguistically-informed computational approach to the reconstruction of protoforms. By reconciling traditional historical linguistics methods with modern statistically-driven techniques to modelling diachronically pertinent states, it offers a pathway towards more accurate and scalable models of language evolution. We release our code at https://github.com/PromiseDodzi/protoform-reconstruction.

**2. Related work**
Early attempts at automated protoform reconstruction were predominantly based on combinatorial techniques, operating under the principle that "the design of the phonological proto-system is best left to human experts" (Kondrak, 2002: 133). These foundational approaches were essentially rule-driven i.e., protoform reconstruction was mainly a process of applying predetermined linguistic constraints and correspondences.

In this light, Durham & Rogers (1969) pioneered the automation of protoform-reflex relationship investigation through phonological constraint matching techniques. Building upon this foundation, Eastlack (1977) proposes a sophisticated algorithm for reconstructing Latin/Ibero-Romance protoforms from Spanish through a systematic identification of phonological rules using a transducing approach. The evolution of rule-guided reconstruction methods reached significant sophistication with Lowe & Mazaudon (1994), who developed a comprehensive bidirectional framework that they referred to as the Reconstruction Engine (RE). This system incorporates not only rule correspondences but also phonotactic information and semantic cues, enabling reconstruction in both directions: from cognate sets to protoforms and from protoforms to cognate sets. This bidirectional capability represented a substantial advancement in the field.

At the turn of the millennium, the field shifted towards statistical and machine learning techniques, implemented within both supervised and unsupervised frameworks. Within supervised learning paradigms, protoform reconstruction was initially framed as a traditional machine learning classification task. Ciobanu & Dinu (2018) employed conditional random fields for protoform reconstruction by systematically labeling each position in daughter language sequences with corresponding protoform tokens. List et al., (2022) on the other hand, use support vector machines trained on trimmed and aligned cognate sets, incorporating sophisticated encoding schemes for contexts that condition sound changes.

Meloni et al., (2021) framed the task as sequence-to-sequence modeling and implemented an encoder-decoder model for Latin reconstruction from a romance languages wordlist. Building on this, recent approaches within the supervised framework have introduced transformer architectures with increasing sophistication. For example, Kim et al., (2023) proposes a Transformer-based architecture, achieving state-of-the-art performance on both Meloni et al., (2021)'s Romance dataset and a Sinitic dataset, while Akavarapu & Bhattacharya (2024) adapt the MSA Transformer, which is originally developed as a protein language model, for cognate prediction and protoform reconstruction. Most recently, Lu et al., (2024) introduce a bidirectional reranking technique that achieves state-of-the-art performance.

On from the days of combinatorial techniques, recent unsupervised approaches on the other hand have relied on methodologies that include ancestral state reconstruction methods, probabilistic modeling, and hybrid neural network architectures. Although not explicitly designed for protoform reconstruction, Jäger & List (2018) apply ancestral state reconstruction methods in what they call "onomasiological reconstruction", focusing on determining reflexes used to express specific conceptual meanings. Their methodology employs three primary techniques: Maximum Parsimony, Minimal Lateral Networks, and Maximum Likelihood, evaluated across three distinct datasets including the Indo-European lexical cognacy database, the Austronesian Basic Vocabulary Database, and the Basic Words of Chinese Dialects. Despite generally modest performance results, their work demonstrates the pertinence of ancestral state reconstruction methods, particularly those incorporating parsimony principles, for determining phonemes present in protoforms.

Bouchard et al., (2007) and Bouchard-Côté et al., (2009, 2013) develop a series of highly influential approaches that frame protoform reconstruction as probabilistic generative task, modeling probabilities of substitutions, insertions, and deletions conditioned on local phonological contexts. Working with Romance (Bouchard et al., 2007; Bouchard-Côté et al., 2009) and Austronesian (Bouchard-Côté et al., 2009, 2013) datasets, they report impressive performance metrics. Their base models incorporate natural phoneme classes and use expectation-maximization algorithms for parameter optimization. Crucially, their modeling approach is guided by initialization with common sound changes and a language model that guides phonotactic plausibility. Despite the impressive results reported, the unavailability of their source code has limited reproducibility.

He et al., (2022) nevertheless manage to extend this probabilistic framework by employing neural networks to model the required edit operations, reporting substantial improvements on an updated version of Meloni et al., (2021)'s dataset. They compare their results against a baseline reimplementation of Bouchard-Côté et al. (2007)'s model and make their implementation publicly available. More importantly, their approach assumes "access to a simple (phoneme-level) bigram language model of the protolanguage" (He et al., 2022: 1638).

**3. Task description**

Protoform reconstruction involves inferring ancestral lexical forms (referred to as protoforms) from which lexical items in modern languages are hypothetically derived. Therefore, given a set of cognate sets i.e., a set of words that have a common ancestral source, the task is to predict these protoforms without relying on any direct access to ancestral data or pre-trained models of the protolanguage. Consequently, contrary to the approach taken by He et al. (2022), we assume no prior access to latent or learned representations of the protolanguage. Instead, we admit the availability of a set of deterministic, language-specific sound change rules, consistent with traditional historical linguistics practice.

The performance of the proposed method is evaluated against a reimplementation of the baseline model introduced by Bouchard-Côté et al. (2007), allowing for a direct comparison in a similar unsupervised setting. The quality of the protoform predictions is assessed using a range of metrics computed with respect to the gold-standard forms.

Let $\Sigma$ denote the set of phonemes as defined by the International Phonetic Alphabet (IPA), and let $\Sigma^*$ represent the set of all finite-length strings over $\Sigma$. Each lexical item is a member of $\Sigma^*$. We consider a collection of cognate sets denoted by $\mathcal{C}$. Each cognate set $c \in \mathcal{C}$ consists of a set of reflexes $\{w_c^l : l \in \mathcal{L}\}$, where $\mathcal{L}$ is the set of modern languages under consideration, and $w_c^l$ represents the reflex of the cognate in language $l$. Each reflex is assumed to have been independently derived from a common ancestral word form $p_c \in \Sigma^*$ via a known set of language-specific phonological transformation rules $\{r^l : l \in \mathcal{L}\}$, which are presumed to be known a priori. Crucially, the ancestral forms i.e., the *protoforms* $\{p^c : c \in \mathcal{C}\}$, are not directly observed. The objective, therefore, is to reconstruct these protoforms based on the observed reflexes $\{w_c^l\}$ and the known transformation rules $\{r^l\}$, thereby reproducing the inference process of a historical linguist.

**3.1. Dataset**

The dataset employed in this study is taken from Ciobanu & Dinu (2014), which comprises of 5,419 cognate sets with reflexes from five Romance languages: Romanian, French, Italian, Spanish, and Portuguese. Each set includes a Latin protoform, representing the common ancestor from which the modern reflexes have evolved. All lexical items are transcribed using the International Phonetic Alphabet (IPA). Instances of missing data are denoted by a hyphen.

To maintain consistency with prior research on unsupervised protoform reconstruction, vowel length markings, indicated by a colon following the vowel, are removed during preprocessing. However, in contrast to He et al., (2022), reflexes from all five languages, including Romanian, are retained in the dataset, to preserve the full linguistic diversity represented in the source material.

## 4. Model

We propose a three-phase framework for protoform reconstruction from cognate sets. The model operates on pre-aligned cognate sets and employs an approach that combines parsimony-based optimization, rule-based transformation modeling, and evolutionary computation.

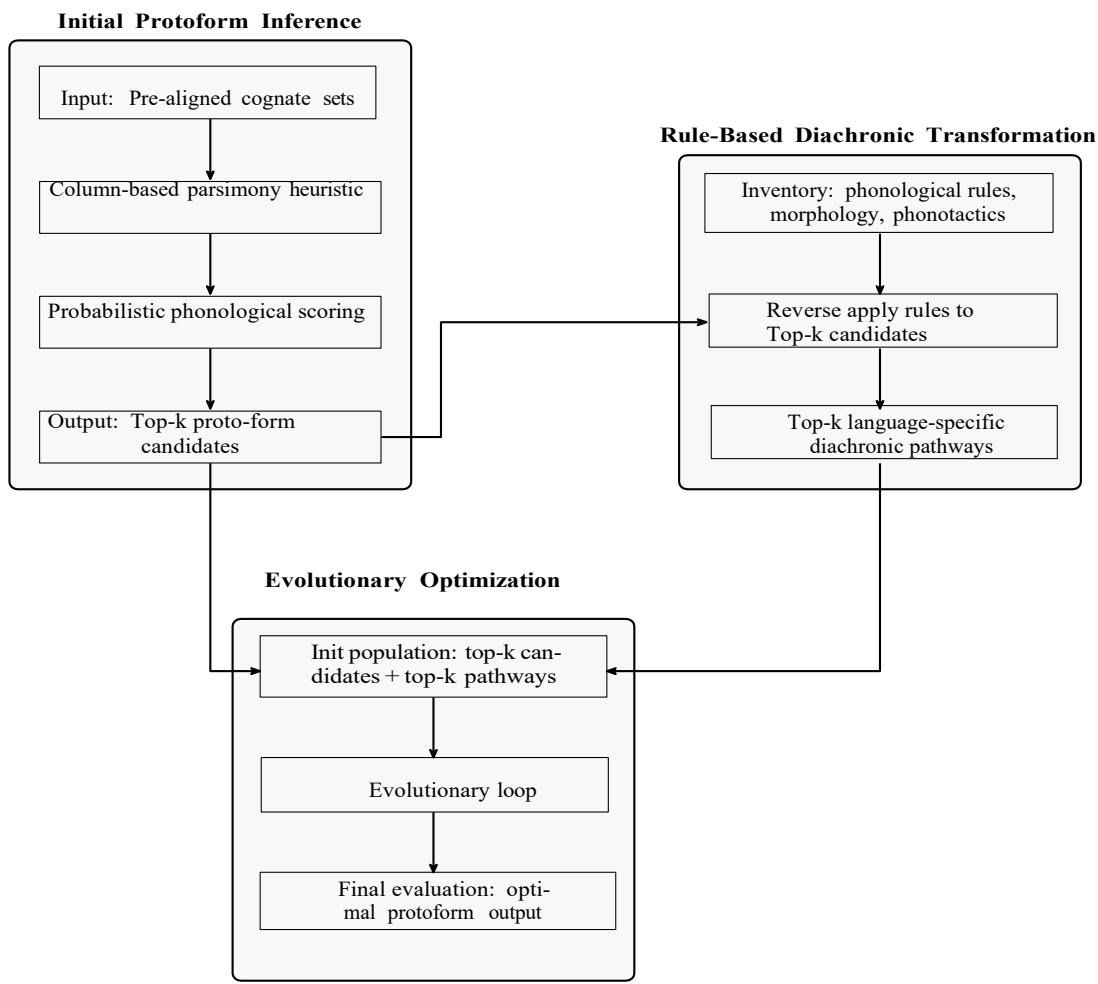

**Figure 2**: *Overview of our protoform reconstruction model, which adopts a hierarchical orientation. The model operates on pre-aligned cognate sets and proceeds through three main stages. Initial inference uses a beam search parsimony algorithm followed by probabilistic phonological scoring to identify high-likelihood reconstructions. This is followed by a rule-guided diachronic transformation which applies a weighted, rule-based system to model language-specific diachronic transformations, generating top-k derivational pathways from highest scoring candidates from initial inference. The final stage involves evolutionary optimization, combining candidates from Phases I and II and iteratively refining them via a fitness-guided evolutionary algorithm to converge on an optimal protoform reconstruction.*

### 4.1. *Phase I: initial protoform inference*

The first phase implements a two-stage protoform reconstruction procedure. The initial stage employs a column-based beam search parsimony algorithm that generates candidate protoforms by minimizing substitution and insertion-deletion costs across alignment columns. We then constrain the search space to prevent combinatorial explosion while at the same time exploring optimal ancestral reconstructions under maximum parsimony criteria.

The second stage applies a probabilistic phonological scoring function to rank the generated candidates returned from the first stage. The scoring metric integrates multiple linguistic factors: context-conditioned phonological change probabilities, edit distance costs, and length-based brevity constraints. This probabilistic framework prioritizes reconstructions that maximize the likelihood of observed sound changes. The integration of this probabilistic phonological evaluation with the parsimony-based generation is to enable us obtain linguistically valid protoform hypotheses.

### 4.2. *Phase II: rule-based diachronic transformation*
The second phase explicitly models rule-governed evolution from reconstructed protoforms. The procedure begins with the compilation of an inventory that includes phonological rules, morphological cues, and phonotactic constraints.

Each transformation rule receives a weight parameter calibrated according to cross-linguistic phonetic naturalness. For the highest ranked candidates returned from phase I, the model reverse applies the weighted rule set to obtain the top-$k$ most probable derivational pathways per language. This produces ranked hypothetical forms that offer optional diachronic pathways that link protoforms to observed reflexes, albeit via the highest ranked candidates from phase I.

### 4.3. *Phase III: Evolutionary Optimization*
The final phase consists of an evolutionary algorithm initialized with two candidate sets: (1) top-k protoforms from the parsimony-based beam search, and (2) language-specific top-k forms derived through weighted reverse transformation rules. The algorithm iteratively evaluates candidates using a composite fitness function that combines likelihood estimates with dynamically adjusted linguistic plausibility constraints. Each round (which can be considered a generation) eliminates low-scoring candidates through selection pressure which mimics natural selection mechanisms.

To maintain diversity and prevent convergence to local optima, the algorithm introduces novel candidate variants through mutation operators when diversity falls below a predetermined threshold. The evolutionary process continues for a maximum number of rounds (generations) or until convergence occurs. Terminal candidates undergo final evaluation to identify the optimal protoform reconstruction that best explains the cognate data under the model's combined criteria.

Figure 2 provides a schematic representation of the model architecture and component interactions. The next subsections detail the algorithmic implementations of each phase.

### 5. Initial protoform inference
In the initial protoform inference phase, we construct a ranked list of protoform candidates $y_c \in \Sigma^*$ for each cognate set $c \in \mathcal{C}$ based solely on the set of attested reflexes $\{w_c^l : l \in$ $\mathcal{L}\}$. We assume that each reflex in a cognate set $w_c^l$ is a transformation of an unobserved protoform candidate $y_c$. The goal is to propose plausible candidates for $y_c$ that minimize divergence from the attested data and conform to phonological expectations. This phase comprises two sub-steps: a column-based parsimony heuristic step, and a phonological-plausibility ranking step.

### 5.1. *Column-based parsimony heuristic*

We thus begin with a character-wise heuristic construction over the aligned reflexes, inspired by traditional parsimony principles. Let the aligned reflexes be represented as a padded matrix $W \in (\Sigma \cup \{-\})^{|\mathcal{L}| \times n}$ where $-$ denotes gap characters introduced for alignment and $n$ is the maximum word length after padding.

Then, for each column $i \in \{1, \ldots, n\}$, we define a character-wise optimization problem. For each character $\sigma \in \Sigma$ occurring in the column $i$ we compute the cost of extending the candidate protoform prefix $y_{<i}$ with $\sigma$ where a gap $-$ incurs a penalty of 1 and a substitution (i.e., $\sigma \neq w_c^l[i]$) also incurs a penalty of 1. Consequently, for a given protoform candidate prefix $y_{<i}$ with cost $C(y_{<i})$, the cost of extension with character $\sigma$ is:

$$C(y_{\leq i}) = C(y_{<i}) + \sum_{l \in \mathcal{L}} \delta(\sigma, w_c^l[i])$$

Where $\delta(\sigma, w_c^l[i]) = 1$ if $w_c^l[i] \neq \sigma$ or $w_c^l[i] = -$, and 0 otherwise.

To avoid combinatorial explosion, we retain only the *top y* protoform candidates at each column based on cumulative cost. The resulting list of candidate protoforms $\{y_c^{(1)}, y_c^{(2)}, \ldots, y_c^{(k)}\}$ provides a heuristic approximation of the most parsimonious protoform candidates. Although this heuristic phase generates structurally minimal protoforms, it does not completely account for phonological realism. We thus introduce a probabilistic ranking scheme that evaluates each protoform candidate $y_c^{(i)}$ against reflexes $\{w_c^l\}$ using a plausibility score $S(y_c^{(i)})$.

### 5.2. *Probabilistic phonological ranking*

Probabilistic phonological ranking of protoform candidates $\{y_c^{(1)}, y_c^{(2)}, \ldots, y_c^{(k)}\}$ is done by first modeling phonological change. This model is used in computing a phonological plausibility score. This score is then adjusted based on auxiliary parameters.

To model change, for each pair of $(y_c^{(i)}, w_c^l)$, we identify all possible edit operations (substitution, insertion, deletion) that transform each generated protoform candidate $y_c^{(i)}$ into $w_c^l$, with each edit operation further classified as:
- Unconditioned: context-free (e.g., /a/ →/e/)
- Conditioned: context-sensitive based on preceding or following segments (e.g., /a/ →/e/ / t_)

Let $m_c^l$ = number of distinct changes between $y_c^{(i)}$ and $w_c^l$, and $cond_c^l$ = number of these changes that are context-sensitive. Then, computing loss as: $loss(y_c^{(i)}) = \sum_{l \in L} 1[y_c^{(i)} \neq w_c^l]$, the phonological plausibility score is defined as a log-scaled score P(D|S), in which $D$ represents the observed divergences and $S$ the structural form of the protoform candidate:

$$P(D \mid S) = m_c \cdot \log_b(h) + \text{cond}_c \cdot \log_b(2) + \sum_{i=\text{loss}+1}^{|L|} [\log_b(i) - \log_b(i - \text{loss})] - |L|$$

Where $m_c = \sum_l m_c^l$, $\text{cond}_c = \sum_l \text{cond}_c^l$, $h$ = homotopy parameter controlling allowable variation (e.g., $h = 10$) and $b$ = complexity base controlling sharpness of penalty (e.g., $b = 10000$). It should be noted that $|L|$ in $p(D|S)$ is to normalize or penalize scores based on the total number of languages considered, ensuring scores are comparable across different cognate sets of varying sizes.

To favor conservative reconstructions, we introduce two auxiliary penalties:

$$P_{\text{brev}}(y_c^{(i)}) = -\lambda \cdot \left| \text{len}(y_c^{(i)}) - \text{avg\_len}(\{w_c^l\}) \right|$$

$$P_{\text{edit}}(y_c^{(i)}) = -\mu \cdot \frac{1}{|L|} \sum_l \text{lev}(y_c^{(i)}, w_c^l)$$

Where $\text{lev}(y_c^{(i)}, w_c^l)$ denotes Levenshtein distance between a protoform candidate $y_c^{(i)}$ and a reflex $w_c^l$ and $\lambda, \mu$ are weights which can be tuned (typically $\lambda = \mu = 5.0$). The final composite score then is:

$$\text{Score}(y_c^{(i)}) = \frac{1}{|L|} \sum_l \left[ P(D \mid S) + P_{\text{brev}}(y_c^{(i)}) + P_{\text{edit}}(y_c^{(i)}) \right]$$

Candidates are subsequently ranked by this score, and the *top* $y_c$ forms are selected for further processing. More importantly, this ranking phase emulates the decision process of historical linguists, by introducing context-sensitive change modeling and phonological plausibility metrics. By jointly minimizing edit cost, rewarding brevity, and enforcing regular phonological transformations, the model proposes protoform candidates that are both cost efficient and linguistically motivated.

## 6. Rule-based transformation

The second phase of our model implements a reverse-engineering approach to protoform reconstruction through weighted rule-based transformations. This phase operates on the principle that the data-driven reconstructions, although benefiting from linguistically conditioned ranking, are constrained in their possible phonemic inventories and that applying language-specific inverse diachronic sound laws to these, offers a fruitful pathway to identifying the most realistic protoform reconstruction. Consequently, given the *top* $y_c$ reconstructions from phase I, the model assumes each language can propose a unique pathway $q_c^{(i)} \in \Sigma^*$ through language-specific sound laws acting on individual members of $top\ y_c$. The goal then is to instantiate these pathways by applying inverse transformations $\mathcal{R}_\ell^{-1}$ on $y_c^{(i)}$.

Nevertheless, the transformation system goes beyond just rules and employs a more comprehensive linguistic inventory comprising:
- The phonological rules: $\mathcal{R}_\ell = (s_i, t_i): s_i \rightarrow t_i$ for each language $l$
- Morphological cues: $\mathcal{M}$ defining valid morphemic boundaries and affixation patterns
- Phonotactic constraints: $\mathcal{P}$ specifying permissible sound sequences and syllable structures

This notwithstanding, phonological rules, more than any other item in the inventory, constitute the transformational base. Consequently, each transformation rule $(s_i, t_i) \in \mathcal{R}_\ell$ receives a naturalness weight $\omega(s_i, t_i)$ calculated as:

$$\omega(s_i, t_i) = \frac{1}{1 + \alpha \cdot |len(s_i) - len(t_i)|} \cdot \phi(s_i, t_i)$$

Where $\alpha$ is a length penalty parameter and $\phi(s_i, t_i)$ represents a phonetic similarity bonus: $\phi(s_i, t_i) = 1.5$ if $s_i, t_i$ belong to the same phonetic class, 1.0 otherwise. Phonetic classes include natural groupings: vowels $\mathcal{V}$, stops $\mathcal{S}$, fricatives $\mathcal{F}$, liquids $\mathcal{L}$, and nasals $\mathcal{N}$

Taking each member of $top\ y_c$, $y_c^{(i)}$, the algorithm generates various pathways through iterative application of inverse rules $\mathcal{R}_\ell^{-1} = (t_i, s_i, \omega(s_i, t_i)): (s_i, t_i) \in \mathcal{R}_\ell$. For each input candidate $y_c^{(i)}$, and language-specific rule $r^l$, we compute the set of candidate reverse transformations $\mathcal{T}_{r^l}^{(\leq d)}\left(y_c^{(i)}\right)$ by recursively applying inverse rules up to a bounded depth $d$. This process explores all substitutions $t \rightarrow s$ from the inverse rule set $\mathcal{R}_\ell^{-1}$, generating new candidates at each recursion step. The resulting candidates are then scored to identify the most plausible forms. To prevent combinatorial explosion, the algorithm maintains at most $Q$ candidate pathways per iteration, retaining only the highest-scoring transformations for each language according to a linguistic motivated composite score:

$$\mathcal{S}\left(q_c^{(i)}\right) = \sum_i \mathcal{M}\_score + \mathcal{P}\_score$$

where $\mathcal{M}\_score$ and $\mathcal{P}\_score$ are based on penalties and bonuses for morphological and phonotactic well-formedness i.e., transformations that contain valid morphological cues and phonotactic structures are rewarded while those that contain illicit phonotactic structures are severely penalized (for e.g., transformations containing three successive consonants are penalized by 0.6 subtraction).

This phase ends with the algorithm generating the $top\ k$ candidate pathways $q_c^{(i)}: i = 1, ..., k$ for each language, hence returning a total of $\sum_{l \in L} top_k, q_c^{(i)}$. These candidate pathways, together with $y_c^{(i)}$ serve as inputs to the subsequent evolutionary optimization phase. The idea is that by obtaining the best data-driven candidates, and rule-driven pathways from these two phases, the evolutionary phase can take advantage of the most cost-efficient and linguistically-valid seeds.

**7. Evolutionary Optimization**

The final phase implements an evolutionary algorithm that optimizes protoform reconstruction by approximating $P(w_c^l: l \in \mathcal{L}|p_c)$ for each cognate set $c \in \mathcal{C}$. The algorithm synthesizes candidates from both the probabilistically ranked parsimony heuristics phase and the rule-based transformation phase, employing iterative selection and mutation to converge on the single optimal protoform $p^c$ that best explains the observed reflexes. Consequently, given a cognate set $c$ with observed reflexes $w_c^l: l \in \mathcal{L}$, the evolutionary algorithm seeks to find: $p^c = arg\ max\ p \in \Sigma^* P(w_c^l: l \in \mathcal{L}|p^c)$, under the assumption that each reflex $w_c^l$ derives from the protoform $p_c$ through language-specific transformation rules $r^l: w_c^l = r^l(p_c) + \epsilon_l$, where $\epsilon$ represents stochastic variation in the diachronic process.

To start, for each cognate set $c \in \mathcal{C}$, the evolutionary algorithm initializes with a seed protoform set $\mathcal{P}_c^{(0)}$ defined as: $P_c^{(0)} = K_{(parsimony)}(c) \cup K_{(transform)}$, where: $K_{(parsimony)}(c) = y_c^1, y_c^2, \dots, y_c^k$; and $K_{(transform)}(c) = q_c^1, q_c^2, \dots, q_c^k$. Thus, the *top k* protoform candidates from phase I and the top language-specific pathways from phase II, are fed as seeds to the evolutionary algorithm. This is to ensure that both parsimony-derived and transformation-derived hypotheses contribute to the evolutionary search space. The core fitness function approximates the likelihood $P(w_c^l|p_c)$ through a composite scoring mechanism: $\mathcal{F}(p_c) = log\ P(w_c^l: l \in \mathcal{L}|p_c) + log\ P(p_c)$. The likelihood term models the probability of observing the reflexes given the protoform:

$$log\ P(w_c^l|p_c) = \sum 1 \in \mathcal{L} \psi_1 \cdot log\ P(w_c l^l|p_c)$$

Where $\psi_1$ represents phylogenetic weights and $P(w_c^l|p_c) = sim(r^l(p_c), w_c^l)$. This latter function is a similarity function that captures both edit distance and phonetic class relationships:

$$sim(x, y) = 1 - \frac{lev(x, y)}{max(|x|, |y|)} + \sum_{i=1}^{min(|x|,|y|)} \phi(x[i], y[i])$$

where $\phi$ again represents phonetic similarity. The prior $P(p_c)$ component on the other hand incorporates morphological and phonotactic well-formedness so that $log\ P(p_c) = \alpha \cdot Smorph(p_c) + \beta \cdot Sphonotactic(p_c)$ where $Smorph(p_c)$ rewards morphologically plausible structures and $Sphonotactic(p_c)$ enforces phonotactic constraints.

The evolutionary process operates through multiple rounds of competitive selection. At each round $t$, the protoform set $\mathcal{P}_c^{(t)}$ undergoes iterative elimination. For $R$ elimination rounds, candidates are ranked by fitness and the lowest-scoring individuals are removed:

$$\mathcal{P}_c^{(t,r+1)} = p p \in \mathcal{P}c^{(t,r)}: \text{rank}\mathcal{F}(p) \leq \left|\mathcal{P}_c^{(t,r)}\right| - \eta_r$$

Where $\eta_r = \max\left(1, \left|\left|\mathcal{P}_c^{(t,r)}\right|/5\right|\right)$ ensures elimination of poor performing candidates per round. Moreover, to prevent extreme length variations, candidates receive adjusted fitness scores. Also, when the diversity of a protoform set falls below threshold $\theta_{div}$, the algorithm introduces variant candidates through three mutation operators:

i. Vowel mutation: $\text{mutate}_V(p_c) = p_c[1:i-1] \circ p_c[i+1:|p_c|]$ where $v' \sim \text{Uniform}(\mathcal{V})$ for position $i$ such that $p_c[i] \in \mathcal{V}$.
ii. Morphological mutation: $\text{mutate}M(p_c) = \text{stem}(p_c) \circ s'$ if $p_c$ has recognizable suffix $s \backslash p_c \circ s'$ with probability $p$
iii. Phonotactic mutation: $\text{mutate}C(p_c) = \text{simplify\_clusters}(p_c, \mathcal{C}valid)$

The evolutionary process for cognate set $c$ terminates under one of three possible conditions: a 'single survivor' condition: $\left|\mathcal{P}_c^{(t)}\right| = 1$; a 'maximum generations' condition: $t \geq T_{max}$; and a 'fitness convergence' condition: no improvement in best fitness for $\delta$ consecutive rounds. Upon termination, the algorithm returns the single best protoform for cognate set. This form represents the algorithm's best approximation of the ancestral form that maximizes $P(w_c^l: l \in \mathcal{L}|p^c)$. The evolutionary optimization phase can thus be assumed to be a synthesizing of the insights from both parsimony-based and rule-based approaches, converging on a single optimal protoform reconstruction that best explains the observed cognate data under the model's probabilistic framework.

## 8. Experiments

### 8.1. *Evaluation criteria*
We adopt a set of standardized evaluation metrics that allow us to measure the validity of the reconstructions proposed by the algorithm vis-à-vis the reference forms. These metrics assess reconstruction quality at the character and phonological feature levels.

The first metric, character accuracy(C_ACC), denotes the percentage of characters in the predicted forms that exactly match the corresponding characters in the correct forms. The other metric that concerns character is a length-normalized Edit Distance(EDIT_DIST), which quantifies the average number of insertions, deletions, or substitutions needed to convert the predicted protoforms into the reference protoforms, based on Levenshtein (1966).

The second metric involves phonological features. The Feature Distance (FEAT_DIST), measures phonological dissimilarity by counting mismatches in articulatory features (like voicing, place/manner of articulation) rather than literal characters, and takes the mean of this. This metric therefore accounts for the relative severity of phoneme substitution errors by quantifying how many phonological features differ between the predicted and target forms. For example, substituting /t/ with /d/ (differing only in voicing) incurs a lower penalty than substituting /k/ with /z/, which differs in multiple features.

In addition, we also calculate the Vowel Error Rate (VER) and Consonant Error Rate (CER), which measure the total vowel and consonant errors, normalized by the total phonemes in the reference forms. Together, these metrics offer a linguistically informed and comprehensive evaluation of model performance.

### 8.2. *Baseline reconstruction*

To establish a meaningful performance benchmark, we compare our model against a reimplementation of the base model originally proposed by Bouchard-Côté et al., (2007), as implemented by He et al., (2022). While our baseline inherits the overall architecture and modeling framework of the original, several modifications are introduced to better suit the scope and constraints of our study. Specifically, we do not adopt the natural phonemic class definitions employed in Bouchard-Côté et al., (2007) - which in any case are not publicly available, opting instead for a simpler division into vowel and consonant sets. Additionally, we follow He et al., (2022) in setting the model's initial probability values ([0.7, 0.1, 0.2]) and smoothing constant (0.1), as the original paper does not explicitly state these. Our dataset also diverges from prior work in that it includes Romanian and omits the complex preprocessing pipelines described in both previous studies.

Despite these differences, the core structure and inference machinery of the model remain intact, preserving its role as a valid comparative baseline. This reimplemented base model thus serves not only as a reference point for evaluating the performance gains of our proposed modifications, but also ensures that improvements are attributable to our model's architectural novelty.

### 8.3. *Ablation studies*

Our model adopts an architecture comprising three principal phases: (i) a probabilistically-ranked heuristic parsimony phase, (ii) a rule-based transformation phase, and (iii) an evolutionary search phase, which also leverages probabilistic heuristics. Let $\mathcal{M} = \mathcal{P}_1, \mathcal{P}_2, \mathcal{P}_3$ represent the full set of components constituting the model, corresponding respectively to the three phases mentioned above.

For purposes of recall, given a set of cognate sets $\mathcal{C} = c_1, c_2, \ldots, c_n$, the model predicts a protoform $p_c \in \Sigma^*$ for each $c \in \mathcal{C}$. In Phase I ($\mathcal{P}_1$), no protoform candidates are pre-supplied; instead, the system generates a probabilistically ranked list of candidate protoforms $y_c^{(i)} \subseteq \Sigma^*$ based on heuristic parsimony across aligned cognate columns. These candidates are assigned an approximated likelihood score and sorted accordingly. The top-$y$ candidates, $Y_c^{(k)}$, are selected for downstream processing. Phase II ($\mathcal{P}_2$) applies language specific transformational phonological rules to $y_c^{(i)}$. This phase produces a set of language-specific pathways $q_c^{(i)} \subseteq \Sigma^*$, where each $q_c$ results from deterministic application of manually encoded transformation rules. Again, a top-$k$ subset $Q_c^{(k)}$ is what is retained based on language-specific scoring heuristics. The union of retained candidates from both previous phases i.e., $Y_c^{(k)} \cup Q_c^{(k)}$, serves as the input to Phase III ($\mathcal{P}_3$). This final phase performs an evolutionary search, mutating and recombining candidate forms when thresholds are not met, to optimize reconstruction accuracy. It outputs a final predicted protoform $p_c \in \Sigma^*$.

Our ablation studies are designed to evaluate the contribution of the individual components of $\mathcal{M}$, and address three key aspects:
  i. The effect of each phase $\mathcal{P}_i$ on overall model performance,
  ii. The influence of probabilistic ranking (specific to $\mathcal{P}_1$), and
  iii. The effect of rule-based transformations on reconstructions.

To isolate the contribution of each phase, we compare the full model $\mathcal{M}$ to variants where individual phases are ablated. Specifically, we evaluate $\mathcal{M}_{-\mathcal{P}_1}$ - Model without Phase I, $\mathcal{M}_{-\mathcal{P}_2}$ - Model without Phase II, $\mathcal{M}_{-\mathcal{P}_3}$ - Model without Phase III. Each variant is benchmarked on the same reconstruction task, and average reconstruction quality is reported using the standard adopted metrics.

To assess the importance of probabilistic ranking in Phase I, we decouple it from the parsimony-based generation process. In this experiment, unranked candidates $y_c^{(i)} \in \Sigma^*$ are passed to Phase III in the order they are generated (i.e., raw parsimony order), bypassing the probabilistic ranking step. Let this variant be denoted $\mathcal{M}$unranked. We compare $\mathcal{M}$unranked against the ranked baseline to quantify the benefit of ranking.

## 9. Results and Discussion

Table 2 shows the performance of different models on the dataset. Except for mean edit distance (EDIT_DIST) and feature distance (FEAT_DIST), our model, *Ranked Prob-Evo*, performs best on all other metrics. *Ranked Prob-Evo-Ext* is a variant of our model in which the seed forms for evolution optimization includes diachronic pathways generated via reverse-transformation rules applied to each reflex of a cognate set i.e., $\mathcal{R}_\ell^{-1} w_c^l$.

| Model | C_ACC | CER | VER | EDIT_DIST | FEAT_DIST |
|---|---|---|---|---|---|
| Ranked Prob-Evo | **50.64** | **22.31** | **26.95** | 4.63 | 2.68 |
| Ranked Prob-Evo-Ext | 50.04 | 22.74 | 27.12 | **4.61** | 2.75 |
| Ranked Path-Prob | 49.88 | 22.64 | 27.38 | 4.62 | 2.77 |
| $\mathcal{M}$ranked | 47.15 | 24.69 | 28.06 | 4.91 | 2.71 |
| $\mathcal{M}$unranked | 47.27 | 24.03 | 28.60 | 4.89 | **2.66** |
| Base model | 34.85 | 34.24 | 30.80 | 5.87 | 3.36 |

**Table 2**: *Average performance of the protoform prediction models, with bold indicating the best-performing model for each metric.*

*Ranked Path-Prob* is a truncated version of our model in which only highest language-specific diachronic pathways derived from highest ranked parsimony candidates, $Q_c^{(k)}$, are fed as seeds for the optimization of the evolution search process. $\mathcal{M}$ranked is the model component that outputs ranked candidates obtained from column-wise parsimony, while $\mathcal{M}$unranked is that part of the model that retains the most parsimonious form obtained after column-wise parsimony heuristics. Finally, the base model is the baseline reimplementation of Bouchard-Côté et al. (2007)'s model.

Our proposed model substantially outperforms the classical baseline and all other variants across all evaluation metrics, with the exception of mean edit distance (EDIT_DIST) and mean feature distance (FEAT_DIST). On these later metrics, *Ranked Prob-Evo-Ext* and $\mathcal{M}$unranked achieve marginally higher score (by 0.02) respectively. However, these models exhibit lower character-level accuracy, higher consonant and vowel error rates - and greater phonological feature distance (FEAT_DIST) for *Ranked Prob-Evo-Ext*; greater mean edit distance (EDIT_DIST) for $\mathcal{M}$unranked.

While for $\mathcal{M}$unranked this is due to the constraint inherent in the phonemic inventory induced by its cost-minimization principles, for *Ranked Prob-Evo-Ext*, the discrepancy arises from the additional transformational pathways made available via $\mathcal{R}_\ell^{-1}$ acting on individual reflexes. Specifically, by incorporating reverse-rule transformations derived both from the reflexes of cognate sets and from reverse transformations of $y_c^{(i)}$, this model gains access to a broader set of rule-transformed pathways during evolution optimization. These transformations, it should be noted, are biased by a naturalness weight $\omega(s_i, t_i)$, which favors phonologically plausible transformations - i.e., those with minimal length discrepancies and within-class substitutions.

Crucially, because this weighting is applied to both the $y_c^{(i)}$-derived and reflex-derived pathways, it effectively doubles the influence of the naturalness prior. As a result, the model's optimization process becomes skewed toward selecting edits that are more "superficially plausible" or low-cost under the "naturalness pressure" via mutation, even if they diverge from the correct phonological target. This bias leads to slightly shorter edit paths on average, thereby improving the edit distance metric, but at the cost of linguistic pertinence, which manifests as reduced character accuracy and increased error rates in phonological features.

Given the facts noted for *Ranked Prob-Evo-Ext* we further investigated the relationship between increasing the number of transformation rules and performance in our model. The results are shown in Figure 3. Across model configurations incorporating between 1 and 16 transformation rules, we observe a modest and non-monotonic improvement in performance metrics up to the 14-rule configuration. Beyond this point, performance becomes inconsistent: reductions in the Vowel Error Rate (VER) are not consistently paralleled by reductions in the Consonant Error Rate (CER), and decreases in Edit Distance (EDIT_DIST) do not reliably correspond to improvements in Phonological Feature Distance (FEAT_DIST). By the 16-rule configuration, however, the model achieves its lowest CER and VER values, yielding the minimal EDIT_DIST, although this does not coincide with the minimum FEAT_DIST.

These results, although not to be taken in absolute terms as other confounding factors can be involved (e.g., the nature and pertinence the rules being introduced), nevertheless suggest that, beyond a certain threshold in rule-set size, there emerges a trade-off in the model's behavior: the model increasingly prioritizes transformations that minimize surface-level discrepancies, favoring shorter and more "natural" edit paths, even when such transformations compromise phonological featural alignment. This supports our earlier analysis that models with richer, heavily weighted rule spaces tend to prioritize "natural" or low-cost edits, even if they violate phonological feature faithfulness. Notably, the 14-rule configuration, which is adopted in the final model, marks a critical inflection point in this trend. At this configuration, FEAT_DIST improves relative to both smaller and larger rule sets, while CER, VER and EDIT_DIST remain competitive. This indicates that at this stage, the rules contribute meaningful linguistic generalizations rather than introducing redundant or degenerate transformations. As a result, the 14-rule model achieves a more optimal trade-off between surface-level edit efficiency and phonological well-formedness, supporting its selection as the preferred threshold.

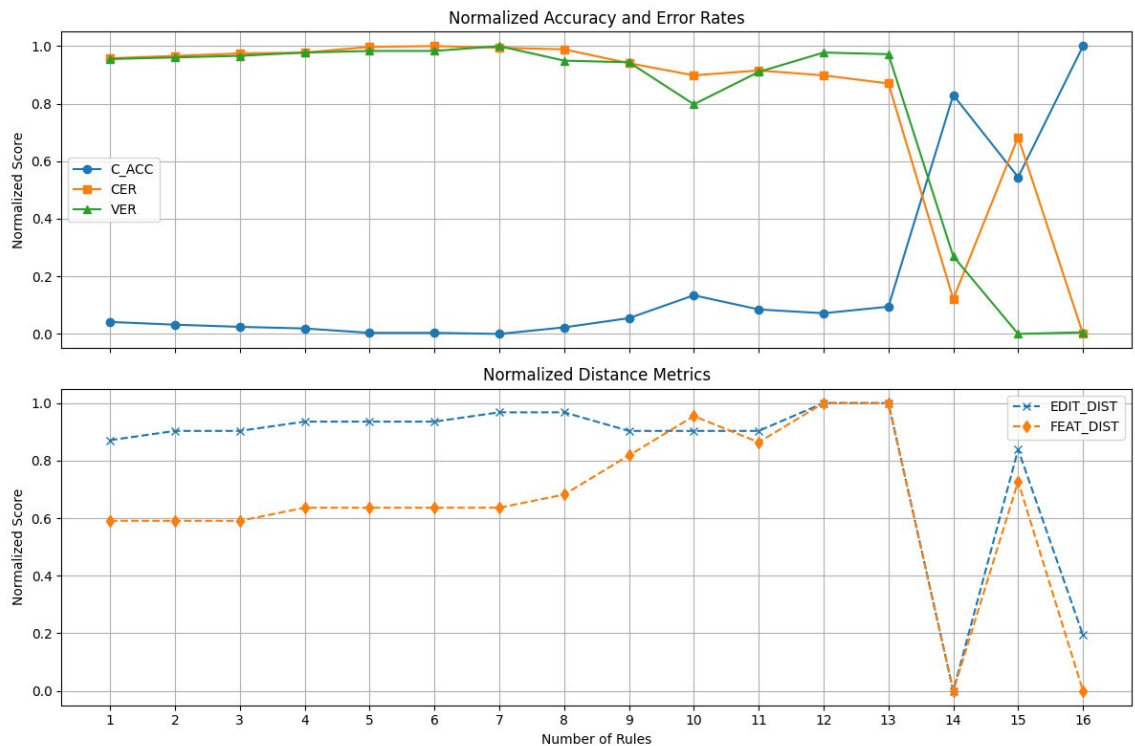

**Figure 3**: *Normalized performance metrics of the protoform prediction model across increasing numbers of rule sets. Character accuracy (C_ACC), consonant error rate (CER), and vowel error rate (VER) are shown in the top panel, while edit distance (EDIT_DIST) and feature distance (FEAT_DIST) are presented in the lower panel. All metrics have been min-max normalized to highlight relative trends. The plots indicate that rule introduction has a marginal but consistent effect on model behavior, with slight trade-offs between surface and phonological accuracy.*

A second point of interest in the results concerns the relative performance of $\mathcal{M}$ranked and $\mathcal{M}$unranked. With the exception of character accuracy (C_ACC) and the vowel error rate (VER) metric, $\mathcal{M}$unranked consistently has superior performance compared to $\mathcal{M}$ranked. This pattern can be attributed to the influence of the probabilistic phonological ranking, which inherently favors phonotactically well-formed protoform candidates. Therefore, applying reverse-rule transformations to $Y_c^{(k)}$ effectively biases the derived pathways towards the most phonotactically plausible protoform candidates for each language, allowing for the evolutionary phase to optimize for both edit-efficiency and phonological well-formedness.

Finally, the results of *Ranked Path-Prob* indicate that the rule-based transformation phase exerts the most significant positive influence on model performance, although as noted earlier, too many of these rule-based transformations introduce biases that manifest as trade-offs. More importantly, the application of transformation rules enables the generation of novel phonemes in the reconstructed protoforms, which are not available in cognate sets, thereby enhancing both the diversity and representational adequacy of candidate reconstructions beyond the limitations of strictly parsimony-based approaches.

Notably, although the transformation phase in isolation ($\mathcal{M}_{-\mathcal{P}_2}$) yields suboptimal performance compared to the complete model, its performance profile diverges meaningfully from that of $\mathcal{M}un$ranked representing $\mathcal{M}_{-\mathcal{P}_1}$. Specifically, *Ranked Path-Prob* representing ($\mathcal{M}_{-\mathcal{P}_2}$) achieves superior character accuracy (C_ACC), consonant error rate (CER), vowel error rate (VER), and mean edit-distance (EDIT_DIST) performance but underperforms in feature-space alignment (FEAT_DIST), whereas $\mathcal{M}un$ranked demonstrates the opposite trend. The subsequent evolutionary search phase in the full model, $\mathcal{M}_{-\mathcal{P}_3}$, takes advantage of this structural variation. The interaction between the different forms from the preceding phases in the evolutionary search process facilitates effective synthesis and mutation, guiding the model toward more accurate protoform reconstructions and ultimately returning improved overall performance.

**10. Conclusion**

In this study, we proposed a hybrid data-driven and rule-guided evolutionary search unsupervised model for protoform reconstruction. While recent approaches in the field predominantly rely on data-driven methods, our work is motivated by the observation that protoforms may contain phonemes absent in reflexes. To address this, our model is designed to emulate key principles of historical linguistic methodology. Evaluated on a dataset of Romance languages, our method demonstrates substantial improvements over existing baselines, suggesting that integrating rule-based constraints into a data-driven framework significantly enhances the modeling of phonological change, which we conceptualize as an evolutionary process.

Looking forward, we envision several avenues for scaling and extending the model. The alignment of reflexes in cognate sets can be optimized using more advanced algorithms, and several components of the current system could benefit from integration with neural architectures to enhance scalability and performance. While our evaluation was limited to five Romance languages, the model is readily extensible to larger datasets and families with deeper phylogenies.

Crucially, the model's design, which allows for the synthesis of rule-guided and data-driven inference, holds particular promise for applications in low-resource language contexts. For linguists working with limited data but possessing knowledge of regular sound correspondences, our framework provides a flexible and interpretable baseline for protoform reconstruction.

**Acknowledgements**
This research is being funded by a grant from the European Research Council (ERC) under the European Union's Horizon Europe Framework Programme (horizon) grant number 101045195.
We wish to thank Abbie Hantgan-Sonko and Fabian Zuk for their feedback on different components of the model. We are also grateful to Andrei Munteanu for sharing code used in the application of the Wordlist Distortion Theory with us; understanding that code was fundamental in the development of our probabilistic phonological ranking component of the model. Finally, we are grateful to members of the BANG team for the support and advice all throughout the development and testing of the model.

# Appendix

## A.1 Reconstruction Limitations

The model demonstrates strong performance on both incomplete and fully observed entries. We do not however investigate the minimum ratio of complete to partial data required to achieve optimal performance. Table 2 presents representative reconstruction examples for data points with complete reflexes available across all languages.

| FRENCH | SPANISH | ITALIAN | ROMANIAN | PORTUGUESE | LATIN (TARGET) | RECONSTRUCTION |
|---|---|---|---|---|---|---|
| -if | -iβo | -ivo | -iw | -ivʊ | -iwʊm | -ivʊm |
| -mã | -mento | -mento | -mint | -meɪntʊ | -mɛntʊm | -mantʊ |
| -ø | -oso | -ozo | -os | -ozʊ | -osʊm | -øsʊm |
| abaj | aβaðia | abbatsia | abatsje | ɐbɐdiɐ | abbatɪam | abatjia |
| abdomɛn | abðomen | addome | abdomen | ɐbdomeɪŋ | abdomɛn | abdomen |
| abɒʁe | aβuriɾ | aborrire | boɾi | ɐvuʁiɪ | abhɔrrɛrɛ | abaʊʁiɾa |
| abʁoʒe | aβroɣaɾ | abrogare | abroga | abʁugaɪ | abrɔgarɛ | abraʊgaa |
| abskɔ̃dʁ | eskondeɾ | askondere | askunde | ʃkuɲdeɪ | abskɔndɛrɛ | askunder |
| absyʁdite | aβsuɾðiðad | assurdita | absurditate | ɐbsuɾədidadi | absʊrdɪtatɛm | absurditada |
| akɑ̃t | akanto | akanto | akantə | ɐkẽntʊ | akantʰʊm | akantʊ |

**Table 1**: IPA transcriptions for some cognate sets from our data after our preprocessing steps, together with gold labels and reconstructions from our unsupervised reconstruction model

## A.2 Hyperparameters

The proposed model operates in three distinct phases, each governed by a specific set of hyperparameters. This section details the key hyperparameter choices for each phase of the model pipeline, explaining the rationale behind their selection and the role they play in ensuring the model's performance and consistency. These hyperparameters were established through preliminary iterative assessment.

### A.2.1 Candidate Generation and Scoring

In the first phase, candidates are generated using beam search with a beam width of 3 and a maximum of 5 iterations. Candidate filtering is initially constrained to the top 10 hypotheses, which is expanded to the top 50 during the reranking stage. The scoring mechanism integrates several weighted components to balance linguistic plausibility with structural consistency. Specifically, brevity and edit distance are each assigned a weight of 5.0, while homotopy is weighted at 10 and structural complexity at 10,000, emphasizing the model's bias toward formally consistent and minimally altered forms.

Various penalty terms are introduced to discourage undesirable patterns. A length mismatch penalty is applied proportionally as 0.3 times the difference in length minus two. Short and long forms incur penalties of −2.0 and −0.4 times the deviation from a 12-character baseline, respectively. Further penalties are imposed for phonological irregularities such as the absence of vowels (−6.0), the presence of invalid sequences (−1.0), and excessive consonant clustering, defined as more than three consecutive consonants (−0.6). Alignment costs are simplified to a uniform value of +1 for both gap insertions and substitutions. Only candidates achieving a similarity score greater than 0.8 are retained for further processing, enforcing a minimum level of fidelity to target phonological patterns.

*A.2.2 Rule-Based Transformation*

The second phase employs rule-based transformations guided by phonotactic constraints, morphological cues, and more importantly, language-specific phonological rules. The phoneme inventory is explicitly partitioned into vowels (*a, e, i, o, u, ʊ, ɛ, ɨ*) and consonants, which span the standard Latin consonantal range i.e., from *b* to *z*, as well as phonemes such as *dʒ* and *ɹ*. Phonological rules applied during this phase are not only deterministic but are also weighted to favor transformations that align with phonetic naturalness. Language relationships are encoded based on a simplified phylogenetic structure, represented as ((French, Spanish), Portuguese, Italian, Romanian), which informs the prioritization and application of transformation rules.

*A.2.3 Evolutionary Optimization*

The final phase consists of the evolutionary optimization procedure. Each round of optimization eliminates the bottom [N / 5] candidates based on scoring metrics, and this process is iterated for up to 20 rounds. To ensure that candidate outputs remain within reasonable structural bounds, a length penalty of 0.8 is applied to outputs deemed too short, while a penalty of 0.5 is assigned to excessively long forms. To prevent convergence to locally optimal but redundant solutions, diversity is maintained by triggering mutations whenever variation among candidates falls below a predefined threshold. The number of candidates evaluated per cognate set is fixed at 10, ensuring consistent statistical sampling across runs.

**A.3 Extended results**

To enable a more comprehensive evaluation of our model's performance, in addition to standard performance metrics, we fell on several supplementary evaluation measures. First, we calculate the Phonotactic Violation Rate (PVR), which quantifies the degree to which the reconstructed outputs conform to established phonotactic constraints of Latin. Second, we define the Normalized Mean Edit Distance (N_EDIT_DIST) as the mean edit distance normalized by the total number of data points. Third, we compute the Feature Error Rate (FER), derived by dividing the Mean Feature Distance (FEAT_DIST) by the total number of evaluated phonological features, providing a normalized measure of feature-level discrepancies. Lastly, the Mean Character Error Rate (MCER), calculated as the total number of character-level errors divided by the number of data points, reflecting the average character error per reconstruction. Extended results are presented in Table 2.

| Metric | Ranked Prob-Evo | Ranked Prob-Evo-Ext | Ranked Path-Prob | $\mathcal{M}$ ranked | $\mathcal{M}$ ranked | Base model |
|---|---|---|---|---|---|---|
| C_ACC | **50.64** | 50.04 | 49.88 | 47.15 | 47.27 | 34.85 |
| MCER | **0.52** | **0.52** | **0.52** | 0.56 | 0.56 | 0.67 |
| CER | **22.31** | 22.74 | 22.64 | 24.69 | 24.03 | 34.24 |
| VER | **26.95** | 27.12 | 27.38 | 28.06 | 28.60 | 30.80 |
| EDIT_DIST | 4.63 | **4.61** | 4.62 | 4.91 | 4.89 | 5.87 |
| N_EDIT_DIST | **0.51** | **0.51** | **0.51** | 0.55 | 0.55 | 0.66 |
| FEAT_DIST | 2.68 | 2.75 | 2.77 | 2.71 | **2.66** | 3.36 |
| FER | **0.06** | **0.06** | **0.06** | **0.06** | **0.06** | 0.08 |
| PVR | 0.19 | **0.17** | 0.18 | **0.17** | 0.19 | 0.16 |

**Table 2**: *Average performance of the protoform prediction models across extended performance metrics, with bold indicating the best-performing model for each metric.*